\documentclass[10pt,twocolumn,letterpaper]{article}

\usepackage[pagenumbers]{cfgs/wacv} %

\usepackage{graphicx}
\usepackage{amsmath}
\usepackage{amssymb}
\usepackage{booktabs}
\usepackage{mathabx}
\usepackage{wasysym}
\usepackage{capt-of}
\usepackage{adjustbox}
\usepackage[dvipsnames]{xcolor}
\usepackage[accsupp]{axessibility}
\usepackage{times}
\usepackage{epsfig}
\usepackage{amssymb}
\usepackage{cuted}

\usepackage[pagebackref,breaklinks,colorlinks]{hyperref}

\usepackage[capitalize]{cleveref}
\crefname{section}{Sec.}{Secs.}
\Crefname{section}{Section}{Sections}
\Crefname{table}{Table}{Tables}
\crefname{table}{Tab.}{Tabs.}

\def\wacvPaperID{441} %
\def\confName{WACV}
\def\confYear{2025}

\usepackage{graphicx}
\usepackage{amsmath}
\usepackage{amssymb}
\usepackage{booktabs}
\usepackage{bbold}
\usepackage{multirow}
\usepackage{bm}
\usepackage{arydshln}
\usepackage{xspace}
\usepackage{url}
\usepackage{comment}
\usepackage{mathabx}
\usepackage{wasysym}
\usepackage{capt-of}
\usepackage{adjustbox}
\usepackage[dvipsnames]{xcolor}
\usepackage{times}
\usepackage{epsfig}
\usepackage{amssymb}
\usepackage{cuted}

\newcommand*{\affaddr}[1]{#1} 
\newcommand*{\affmark}[1][*]{\textsuperscript{#1}}

\definecolor{GreenColor}{rgb}{0.137,0.573,0.565}
\definecolor{OrangeColor}{rgb}{0.914,0.541,0.0.141}
\definecolor{PurpleColor}{rgb}{0.5,0,0.7}
\definecolor{BlueColor}{rgb}{0,0.725,0.949}
\definecolor{PinkColor}{rgb}{0.9843,0.19215,0.6}

\newcommand\blfootnote[1]{%
  \begingroup
  \renewcommand\thefootnote{}\footnote{#1}%
  \addtocounter{footnote}{-1}%
  \endgroup
}

\makeatletter
\newcommand{\figcaption}[1]{\def\@captype{figure}\caption{#1}}
\newcommand{\tblcaption}[1]{\def\@captype{table}\caption{#1}}
\newcommand{\customparagraph}[1]{\par{\noindent\textbf{#1:}}}
\newcommand{\textcite}[1]{``\textit{#1}''}

\makeatother

\DeclareRobustCommand\red{\textcolor{red}}
\DeclareRobustCommand\blue{\textcolor{blue}}
\DeclareRobustCommand\green{\color[rgb]{0,0.5,0}}

\def\tbf{\textbf}

\def\chi{Proceedings of the SIGCHI Conference on Human Factors in Computing Systems (CHI)}

\makeatletter
\DeclareRobustCommand\onedot{\futurelet\@let@token\@onedot}
\def\@onedot{\ifx\@let@token.\else.\null\fi\xspace}
\def\eg{\emph{e.g}\onedot} 
\def\ie{\emph{i.e}\onedot} 
 
 \def\vs{\emph{vs}\onedot}
\def\wrt{w.r.t\onedot} 
\def\etal{\emph{et al}\onedot}

\makeatother
\def\egocap{EgoYC2}
\def\nvideo{226}
\def\nuser{44}
\def\vhour{43}
\def\dataratio{11.3\%}
\def\ego{\textit{ego}}
\def\egolike{\textit{ego-like}}
\def\exo{\textit{exo}}

\newcommand{\eccvadd}[1]{#1}
\newcommand{\eccvradd}[1]{#1}
\newcommand{\wacvadd}[1]{#1}

\begin{document}

\title{
Exo2EgoDVC: Dense Video Captioning of Egocentric Procedural Activities Using Web Instructional Videos
}

\author{Takehiko Ohkawa\affmark[1,2], Takuma Yagi\affmark[3], 
Taichi Nishimura\affmark[4], Ryosuke Furuta\affmark[1],\\
Atsushi Hashimoto\affmark[2], Yoshitaka Ushiku\affmark[2], and Yoichi Sato\affmark[1]\vspace{2mm}\\ 
\affaddr{
\affmark[1]The University of Tokyo\ \  
\affmark[2]OMRON SINIC X Corp.\\
\affmark[3]National Institute of Advanced Industrial Science and Technology (AIST)\ \
\affmark[4]LY Corporation
}
\vspace{-35pt}
}
\maketitle

\begin{strip}
\centering
\includegraphics[width=0.9\hsize]{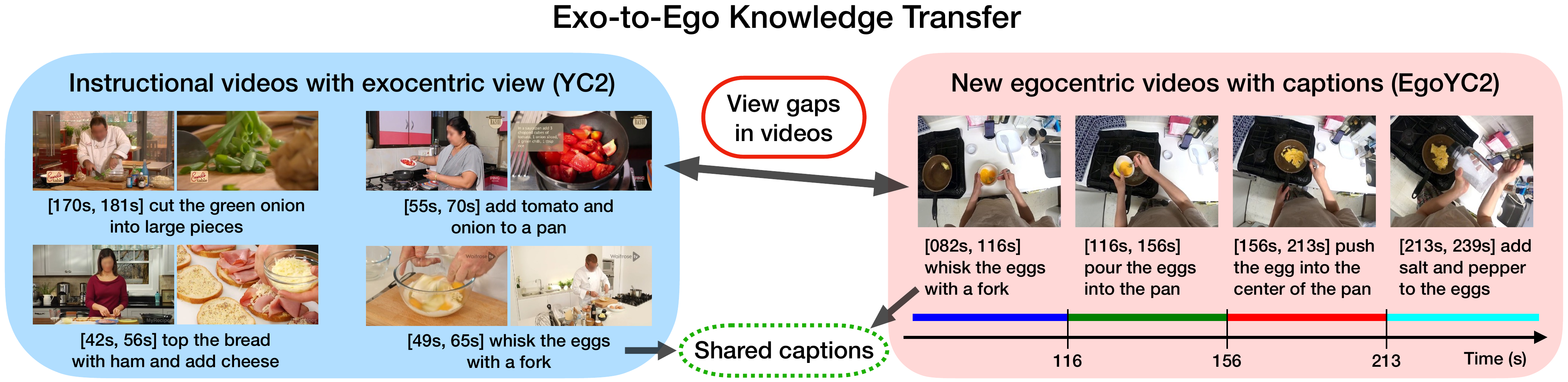}
\captionof{figure}{
\tbf{Our cross-view knowledge transfer of dense video captioning.}
We propose to utilize existing web instructional videos with exocentric views, YouCook2 (YC2)~\cite{zhou:aaai18}, to improve dense video captioning on newly recorded egocentric videos (EgoYC2).
The EgoYC2's captions are annotated by following YC2, enabling the study of transfer learning under view gaps in videos. 
}
\label{fig:teaser}
\end{strip}

\begin{abstract}
We propose a novel benchmark for cross-view knowledge transfer of dense video captioning, adapting models from web instructional videos with exocentric views to an egocentric view.
While dense video captioning (predicting time segments and their captions) is primarily studied with exocentric videos (\eg, YouCook2), benchmarks with egocentric videos are restricted due to data scarcity.
To overcome the limited video availability, transferring knowledge from abundant exocentric web videos is demanded as a practical approach.
However, learning the correspondence between exocentric and egocentric views is difficult due to their dynamic view changes.
The web videos contain shots showing either full-body or hand regions, while the egocentric view is constantly shifting.
This necessitates the in-depth study of cross-view transfer under complex view changes.
To this end, we first create a real-life egocentric dataset (EgoYC2) whose captions follow the definition of YouCook2 captions, enabling transfer learning between these datasets with access to their ground-truth.
To bridge the view gaps, we propose a view-invariant learning method using adversarial training, which consists of pre-training and fine-tuning stages.
Our experiments confirm the effectiveness of overcoming the view change problem and knowledge transfer to egocentric views.
Our benchmark pushes the study of cross-view transfer into a new task domain of dense video captioning and envisions methodologies that describe egocentric videos in natural language.

\end{abstract}

\section{Introduction}
Perceiving procedural human activities from an egocentric (first-person) view has been a long-standing problem.
Compared to action recognition focusing on labeling specific activities~\cite{damen:ijcv21,sener:cvpr22}, video-to-text description extends this realm, offering a detailed textual interpretation of ongoing activities.
This not only facilitates understanding of the task procedure, but enriches intuitive and communicative interfaces between humans and assistive machine systems, \eg, augmented reality~\cite{majil:sensors22} and human-robot interactions~\cite{kang:ijsr23}.

One formulation of video-to-text description is dense video captioning~\cite{krishna:iccv17}, which densely detects time segments of a video and generates their captions.
This task has been studied with instructional videos (\eg, YouCook2~\cite{zhou:aaai18}).
While these instructional videos primarily feature exocentric (third-person) views available on the Web,
egocentric benchmarks for dense video captioning remain underexplored due to limited dataset scale (\eg, 16~hours of MMAC Captions (ego)~\cite{nakamura:mm21}  \vs 176~hours of YouCook2).

Given this data scarcity, finding ways to bridge exocentric (source) and egocentric (target) views is vital to utilizing numerous web exocentric videos to enhance the understanding of egocentric activities.
Prior works of this cross-view transfer have been studied in action recognition~\cite{li:cvpr21,wang:iccv23} and human/hand pose estimation~\cite{khirodkar:iccv23,ohkawa:cvpr23}.

Unlike short-term modeling in these prior works (\eg, estimation of per-frame poses and per-clip actions), dense video captioning requires \eccvradd{addressing longer sequence inputs to model the coherence of each step description}.
This emphasizes the problem of dynamic camera view changes in learning the correspondence between both views.
The source instructional videos are not purely captured from a single fixed camera, but composed of multiple alternating views (\eg, face and hand shots~\cite{miura:tis05}).
In contrast, egocentric videos inherently include dynamic view changes due to the camera wearer's motion, which obstructs learning the procedure~\cite{bansal:eccv22}.
This necessitates adapting captioning models from mixed source views to a moving target view.

In this work, we propose a knowledge transfer benchmark for dense video captioning 
\eccvradd{from web instructional videos with exocentric views to egocentric videos},
with access to their ground-truth during training (\cref{fig:teaser}).
To study this cross-view transfer, we first create a new egocentric dataset, EgoYC2, with caption annotations following a source dataset, YouCook2.
We collect \nvideo{} egocentric cooking videos from \nuser{} users, featuring real-life home kitchens (\vs a laboratory kitchen in MMAC Captions).
These paired datasets reduce discrepancies in caption content and granularity, allowing us to pre-train a model on the source data and fine-tune it on the target data.

To address the view gaps, we encourage pre-training and fine-tuning to be less affected by view-dependent bias, using adversarial invariant feature learning~\cite{ganin:icml15,chen:iccv19}.
The pre-training aims to learn features invariant to the two views in the source data: face shots showing body actions and hand shots focusing on hand-object interactions.
The view-invariant fine-tuning is performed using the source and target datasets, further adapting to the egocentric domain.
\eccvadd{
Additionally, we observe camera motion in egocentric videos intensifies the view gaps; 
}
thus we stabilize the videos by a fine and temporally coherent tracking of hand-object interactions, which consists of hand detection and tracking~\cite{shan:cvpr20,bewley:icip16} and hand-object segmentation~\cite{zhang:eccv22}.

We evaluate our transfer learning method regarding how effectively the method overcomes view changes and efficiently transfers knowledge to the egocentric domain.
Our pre-training with the decoupling of mixed source views has shown further improvement in egocentric video captioning against a naive pre-trained model.
The stabilization of the target view movement with hand tracking improves fine-tuning performance, and additional support of hand-object features is more effective.
Our view-invariant fine-tuning further improves adapting the pre-trained model to the egocentric domain.
Our benchmark allows us to provide a practical solution for transfer learning from exocentric to egocentric videos under dynamic view changes.

Our contributions are summarized as follows:
\begin{itemize}
    \vspace{-9pt}
    \item We offer a new real-life egocentric video dataset (EgoYC2) for dense video captioning, whose captions follow those of exocentric videos (YouCook2).
    \item We propose view-invariant learning in pre-training and fine-tuning with unified adversarial training and video processing that mitigates the view change effects.
    \item We demonstrate how effectively the proposed method overcomes the problem of view changes and efficiently transfers the knowledge to the egocentric domain.
\end{itemize}

\section{Related Work}
\customparagraph{Dense video captioning} 
consists of two sub-tasks localizing multiple time segments occurring in a video and describing their captions.
While Krishna~\etal initially proposes to describe coarse activities~\cite{krishna:iccv17}, subsequent works focus on more fine-grained activities using instructional videos, such as cooking~\cite{zhou:aaai18,nakamura:mm21} (\eg, YouCook2 (YC2)), makeup~\cite{wang:emnlp19}, and daily activity videos~\cite{alayrac:cvpr16,wang:emnlp19,miech:iccv19}.
These instructional videos promote the understanding of the \textit{procedure}~\cite{zhou:aaai18}, a series of steps to accomplish certain tasks.

However, dense video captioning for egocentric (first-person) videos has been less studied compared to exocentric (third-person) benchmarks~\cite{alayrac:cvpr16,zhou:aaai18,wang:emnlp19,tang:cvpr19,miech:iccv19}.
A video captioning dataset for egocentric cooking videos, MMAC Captions~\cite{nakamura:mm21}, contains x11 smaller amount of videos than YC2.
Other related egocentric datasets (\eg, EPIC-KITCHENS~\cite{damen:ijcv21} and Ego4D~\cite{grauman:cvpr22}) are collected with spoken narrations.
Their transcribed texts help define action labels and video-language pre-training~\cite{lin:neurips22,pramanick:iccv23}, but they drastically differ from \textit{procedural} captions~\cite{yang:nips23,yang:cvpr23}.
In addition, its supervision is typically weak and noisy, including incorrect visual grounding, irrelevant caption content, and errors in automatic speech recognition~\cite{miech:cvpr20,ko:cvpr22,soucek:cvpr22,lin:cvpr22}.
{Xu~\etal~\cite{xu:cvpr24} propose to use LLM to refine the noisy captions, enabling egocentric captioning with retrieval.}

We newly construct an egocentric video dataset (EgoYC2) with \textit{procedural} caption annotations. 
Compared to the MMAC Captions dataset captured in a laboratory kitchen, we record real-life cooking activities and annotate captions following the exocentric dataset YC2.
This allows us to let models resolve the view gaps only without considering the gaps in caption content and granularity.
In fact, we use the same vocabulary list as the YC2 and have close average step sizes per video (7.7 in YC2 \vs 6.5 in EgoYC2 \vs 30.1 in MMAC Captions).

\customparagraph{Egocentric perception using exocentric videos}
Exocentric view data can inform the state of humans, actions, and the surrounding environment that are not always observable from a limited 
field-of-view of an egocentric camera.
Thus, knowledge transfer from exocentric to egocentric views is essential to complement egocentric perception, \eg, in action recognition~\cite{sigurdsson:cvpr18,li:cvpr21,truong:arxiv23,wang:iccv23,xue:arxiv23} and human/hand pose estimation~\cite{khirodkar:iccv23,ohkawa:eccv22,ohkawa:cvpr23}.
This transfer has been categorized into two settings: \textit{paired} \vs \textit{unpaired} scenarios.

The \textit{paired} setting assumes that the same actions are captured from different views~\cite{sigurdsson:cvpr18} or using synchronized egocentric and exocentric cameras~\cite{kwon:iccv21,sener:cvpr22,grauman:cvpr24}.
Sigurdsson~\etal propose to learn a shared feature space between both views using metric learning~\cite{sigurdsson:cvpr18}.
With camera calibration and head-camera tracking, other works~\cite{ohkawa:cvpr23,kwon:iccv21,khirodkar:iccv23} project annotated hand-object poses from exocentric to egocentric views.
The \textit{unpaired} setting relaxes this assumption of view correspondence; thus, two videos are neither synchronized nor captured in the same environment.
Several ways to learn view-invariant features in unpaired videos have been proposed, such as knowledge distillation~\cite{li:cvpr21} and cross-view feature alignment~\cite{wang:iccv23,xue:arxiv23} or attention~\cite{truong:arxiv23}.
Other methods employ domain adaptation techniques, such as adversarial training~\cite{choi:wacv20,munro:cvpr20} and pseudo-labeling~\cite{ohkawa:eccv22,ohkawa:access21}.

Our work addresses this cross-view transfer problem for a new task domain, dense video captioning.
Unlike action recognition, the transfer of dense video captioning requires 
\eccvradd{handling longer sequence inputs, which highlights view changes as a longer sequence includes various views}.
Since our work is based on asynchronous videos, we propose an adversarial training method that works in the \textit{unpaired} setting.
\eccvadd{%
Compared to adversarial adaptation in action recognition directly bridging between two datasets~\cite{chen:iccv19,choi:wacv20,munro:cvpr20}, we follow the idea of gradual domain adaptation~\cite{kumar:icml20,liu:neurocomp23,wang:icml22} by splitting a large domain gap caused by different recording setups and distinct viewpoints into smaller gaps, and resolve the gaps step-by-step.
We offer a practical application of gradual adaptation beyond conventional setups~\cite{kumar:icml20,liu:neurocomp23,wang:icml22}, \eg, digit and portrait data.
}

\section{Exo-to-Ego Transfer Learning}
We propose a solution for a transfer learning task of dense video captioning from web instructional videos to egocentric videos, \ie, YouCook2 (source) $\rightarrow$ EgoYC2 (target).
\eccvadd{%
Given the problem of the limited data scale for egocentric video captioning, our proposed method is motivated to utilize external video resources collected on the Web.
We assume that web instructional videos with (dense) captions covering activity classes of egocentric videos are available, and
} 
the target ground-truth is accessible, which follows supervised domain adaptation~\cite{tzeng:iccv15,motiian:iccv17}.
We first introduce our setup and transfer learning method in \cref{sec:viewproc,sec:viewinv}, then describe our model with respect to the representation of the hand-object in \cref{sec:hofeat} and a captioning network in \cref{sec:videocap}.

The challenge of this task is to address the dynamics of view changes in both datasets.
Unlike cross-view transfer in action recognition~\cite{sigurdsson:cvpr18,li:cvpr21,truong:arxiv23,wang:iccv23,xue:arxiv23}, dense video captioning requires modeling
\eccvradd{longer video sequences}, which further complicates the view change problem.
We observe that the source web videos can be decomposed into several shots captured from different views, based on video composition analysis, \eg, instruction classes in how-to videos~\cite{yang:chi23}, scene cut categorization~\cite{pardo:eccv22,miura:tis05}, temporal shot segmentation~\cite{chen2:cvpr21,wu:cvpr23}.
In contrast, egocentric videos are typically untrimmed and captured from a single camera, but involve dynamic scene changes due to the head movement.
This dynamic movement prevents learning the procedure (key steps)~\cite{bansal:eccv22}.
Thus, given the problems of the mixed and moving views, it is necessary to overcome view-dependent bias by adapting models to the target egocentric domain.

\begin{figure*}[t]
\centering
\includegraphics[width=0.8\hsize]{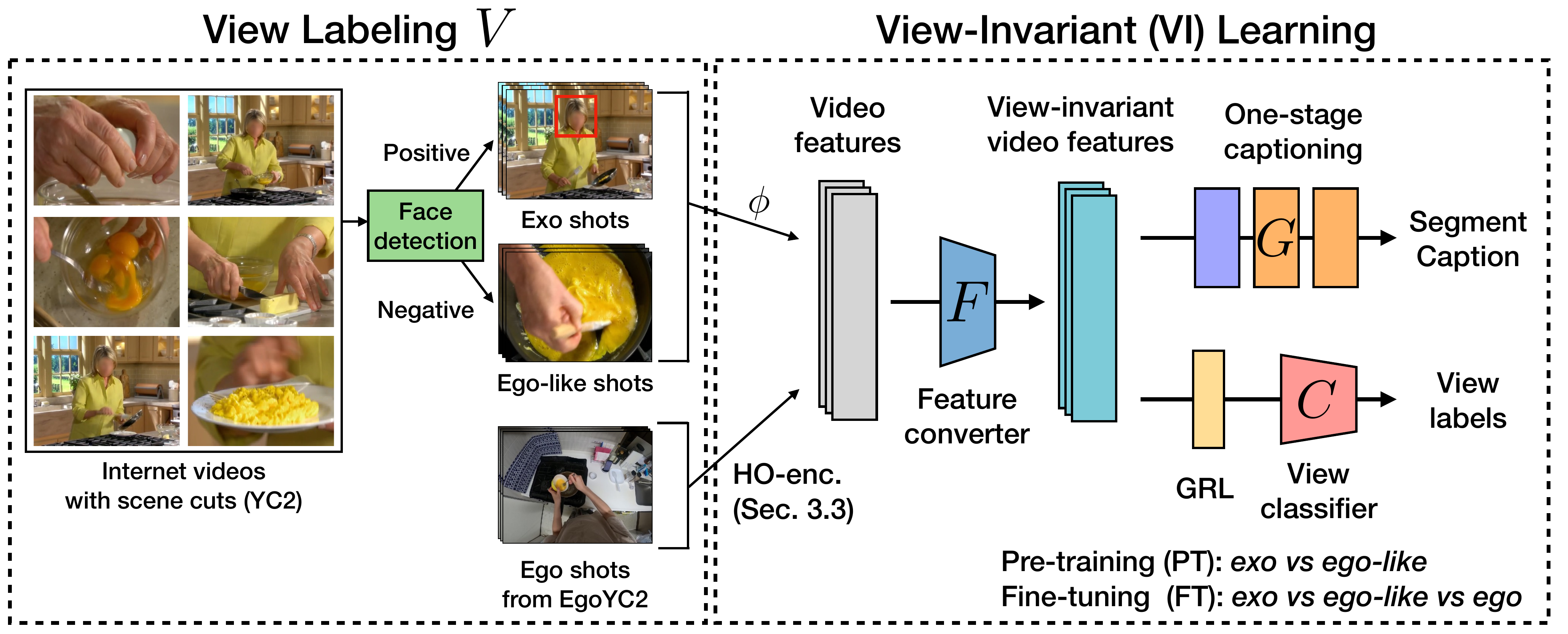}
\caption{
\tbf{View-invariant learning across exocentric and egocentric views.}
(i) We define an intermediate view (\egolike{}) in the source domain, which represents the one between \exo{} and \ego{} views.
We treat source images where the face is detected as the \exo{} view and the others as the \egolike{} view due to its similarity to the \ego{} view. 
We generate video features using a fixed encoder $\phi$ and describe this processing for egocentric videos in \cref{sec:hofeat}.
(ii) We design our view-invariant (VI) learning to gradually adapt from \exo{} to \ego{} views. 
Our method consists of pre-training (PT) on the source data and fine-tuning (FT) across the source and target data. 
Following adversarial domain adaptation~\cite{ganin:icml15}, we train a feature converter $F$ and a view classifier $C$ with a gradient reversal layer (GRL).
This encourages feature learning invariant to the view classes to be classified by $C$.
The former PT takes the source data with the \exo{} and \egolike{} classes, while the latter FT takes all views to align them.
}
\label{fig:method}
\end{figure*}

\customparagraph{Preliminary}
We have access to both labeled source and target datasets $\mathcal{D}_\mathrm{s}$ and $\mathcal{D}_\mathrm{t}$.
These datasets contain
{labeled videos with the size of $m$ ($n$)}, where the input $\mathbf{X}$ is features for a video and $\mathbf{y}$ corresponds to its ground-truth of dense video captions, written as
$\mathcal{D}_\mathrm{s}=\left\{\left(\mathbf{X}_{\mathrm{s}1}, \mathbf{y}_{\mathrm{s}1}\right), \left(\mathbf{X}_{\mathrm{s}2}, \mathbf{y}_{\mathrm{s}2}\right), \ldots,\left(\mathbf{X}_{\mathrm{s}m}, \mathbf{y}_{\mathrm{s}m}\right)\right\}$ 
and 
$\mathcal{D}_\mathrm{t}=\left\{\left(\mathbf{X}_{\mathrm{t}1}, \mathbf{y}_{\mathrm{t}1}\right),\left(\mathbf{X}_{\mathrm{t}2}, \mathbf{y}_{\mathrm{t}2}\right), \ldots,\left(\mathbf{X}_{\mathrm{t}n}, \mathbf{y}_{\mathrm{t}n}\right)\right\}$.
The video features $\mathbf{X}$ are encoded by a fixed feature extractor $\phi$ and represented as a set of frame-wise features: 
{$\mathbf{X}=\left\{\mathbf{x}_1, \mathbf{x}_2, \ldots, \mathbf{x}_{\mathrm{T}}\right\}$}, where $\mathbf{x}$ is frame features and 
{$T$ is a fixed length of a video}.

Unlike a standard formulation of classification tasks~\cite{chen:iccv19,kim:iccv21,munro:cvpr20,choi:wacv20}, dense video captioning poses a complex form of architecture and loss function~\cite{wang:iccv21,zhou:cvpr18,wang:tcsvt21}.
{Here we encapsulate the captioning model $G$ and task loss $\mathcal{L}_{\text{task}}$, and describe their details in Sec.~\ref{sec:videocap}.}
Our inference model consists of a feature converter network $F$ and a task network $G$, learning the mapping from $\mathbf{X}$ to $\mathbf{y}$.
We also define the view class for each frame with a view labeling function $V(\cdot)$, which takes a frame $\mathbf{x}$ and assigns it to one of the predefined view labels (\ie, \exo{}, \egolike{}, and \ego{} as indicated in \cref{sec:viewproc}).
Given an entire model $G \circ F$ and a dataset $\mathcal{D}$, we denote a task loss function for dense video captioning as $\mathcal{L}_{\text{task}}\left(F, G, \mathcal{D}\right)$.
Using the source and target datasets, the joint training on the two datasets is defined as
\begin{equation}
\mathcal{L}_{\text{task}}\left(F, G, \mathcal{D}_{\mathrm{t}}\right)+
\lambda_{\text{src}} \mathcal{L}_{\text{task}}\left(F, G, \mathcal{D}_{\mathrm{s}}\right),
\label{eq:joint}
\end{equation}
where $\lambda_{\text{src}}$ is a weight controlling the training on $\mathcal{D}_{\mathrm{s}}$.

\subsection{View labeling and preprocessing}\label{sec:viewproc}
We construct the view labeling function $V(\cdot)$.
Based on the composition analysis of exocentric cooking videos~\cite{miura:tis05},
the source data can be divided into face shots showing full-body actions and hand shots indicating the necessary attention to specific objects.
These shots are interleaved even in several seconds and changing throughout the video sequence.
Observing the visual similarity between the hand shots and the egocentric images, we define three view classes as \exo{} (face shots) and \egolike{} (hand shots) from the source exocentric videos, and \ego{} views from the target egocentric videos (see the left of \cref{fig:method}).

For the source data, we use face tracking that classifies the face and hand shots, consisting of a face detector~\cite{schroff:cvpr15} and a tracking method SORT~\cite{bewley:icip16}.
We assign positive images with face to the \exo{} view and the rest to the \egolike{} view.
\eccvradd{
Compared to classifier-based scene categorization in Ego-Exo~\cite{li:cvpr21}, our classification based on local facial features is more versatile.
Li~\etal~\cite{li:cvpr21} employ an ego-exo scene classifier trained on Charades-Ego~\cite{sigurdsson:cvpr18} to assign soft labels between \ego{} and \exo{} views.
This scene classifier may exhibit bias due to its training on mostly side views in \cite{sigurdsson:cvpr18}'s \exo{} views, over the front views prevalent in YC2's \exo{} views.
Such biases could limit its generalization to unseen videos like EgoYC2 and YC2.
In contrast, our face detection-based approach can effectively handle various face angles, regardless of side and front views.
}

To reduce the moving impact of the \ego{} view, we propose using a fine-grained and temporally coherent tracking of hand-object interactions, informing actions occurring in moving scenes.
We implement this tracking by using the combination of hand detection-based tracking and frame-wise hand-object segmentation.
Specifically, we track the location of bounding boxes covering two hands using a pre-trained hand detector~\cite{shan:cvpr20,tango:eccvw22} and 
{the SORT algorithm~\cite{bewley:icip16}}.
Additionally, we use the pixel-level location of hands and interacting objects~\cite{zhang:eccv22}, which can provide more precise descriptions of actions.
These two techniques complement each other; the hand tracking is temporally coherent but coarse localization and the segmentation is fine-grained but frame-wise localization
(see \cref{sec:hofeat} for details).

\subsection{Transfer learning via view-invariant learning}\label{sec:viewinv}
We aim to learn view-invariant features among the mixed source views and the unique egocentric view. %
We perform pre-training and fine-tuning separately to handle a larger domain gap.
\eccvadd{
Compared to prior action domain adaptation, our transfer learning will suffer from a larger domain gap as the source and target data are neither constructed within the same dataset~\cite{munro:cvpr20} nor captured from similar viewpoints~\cite{chen:iccv19}.
Here we employ a ``divide-and-conquer'' approach by breaking down a large domain gap into smaller gaps and adapting between the smaller gaps step-by-step, dubbed \textit{gradual} domain adaptation~\cite{kumar:icml20,liu:neurocomp23,wang:icml22}.
}
Specifically, we introduce an intermediate domain in the source data that shares visual similarity with the target data.
This encourages us to resolve the large gap \textit{gradually} with separated training stages.

We facilitate the learning of invariant features with adversarial training, while gradually adapting from \exo{} to \egolike{} and finally to \ego{} view.
\cref{fig:method} illustrates the overview of the view labeling and our learning method.
The video features are converted to learnable representation by the converter $F$. 
These features are fed to the task network $G$ to solve the captioning task and the classifier $C$ to let the converted features be view-independent.
The classifier $C$ is trained by adversarial adaptation with a gradient reversal layer~\cite{ganin:icml15}.
The converter $F$ attempts to produce features undistinguished by the classifier while the classifier is trained to classify their views.

\customparagraph{View-invariant pre-training in source domain}
We pre-train the model on the source data to let the feature converter $F$ produce view-invariant features.
This learning is facilitated by adversarial loss~\cite{ganin:icml15} with the classifier $C$, which learns the mapping from a frame $\mathbf{x}$ to a view label $V(\mathbf{x})$:
\begin{equation}
\mathcal{L}_{\text{adv}}\left(F, C, \mathcal{D}, V\right) 
= \mathbb{E}_{\mathbf{x} \sim X \sim \mathcal{D}} \mathcal{L}_{\text{ce}}\left(
C\left(F\left(\mathbf{x}\right)\right), V\left(\mathbf{x}\right)\right).
\label{eq:adv}
\end{equation}
We use a cross-entropy loss $\mathcal{L}_{\text{ce}}$ as the objective to reuse $\mathcal{L}_{\text{adv}}$ in the three-view classification problem in the next section.
The total loss of the pre-training is defined as 
\begin{equation}
\begin{split}
\mathcal{L}_{\text{task}}\left(F, G, \mathcal{D}_{\mathrm{s}}\right)
-\lambda_{\text{adv}} \mathcal{L}_{\text{adv}}\left(F, C, \mathcal{D}_{\mathrm{s}}, V\right),
\label{eq:vi-pre}
\end{split}
\end{equation}
where $\lambda_{\text{adv}}$ is a controlling weight for the adversarial loss.

\customparagraph{View-invariant fine-tuning across three views}
We follow similar view-invariant learning during the training with the source and target data.
We first initialize the entire model $G \circ F$ from a pre-trained checkpoint (\cref{eq:vi-pre}) and reinitialize the classifier $C$.
Then, we jointly train the model on the source and target data (\cref{eq:joint}), while the classifier $C$ is trained to classify a frame $\mathbf{x}$ from both datasets into the three view classes as
\begin{equation}
\begin{split}
\mathcal{L}_{\text{task}}\left(F, G, \mathcal{D}_{\mathrm{t}}\right)
&+\lambda_{\text{src}} \mathcal{L}_{\text{task}}\left(F, G, \mathcal{D}_{\mathrm{s}}\right) \\
&-\lambda_{\text{adv}} \mathcal{L}_{\text{adv}}\left(F, C, \right\{\mathcal{D}_{\mathrm{s}}, \mathcal{D}_{\mathrm{t}}\left\}, V\right).
\label{eq:vi-train}
\end{split}
\end{equation}

Note that the source web data are typically much larger than recorded egocentric videos ($m \gg n$).
To address this, we employ an undersampling technique for task loss of \cref{eq:joint}, where the number of input source data is balanced to the number of target data per iteration.
Similarly, we address the imbalance of view classes in the adversarial loss of \cref{eq:adv} with the undersampling.

\begin{figure}[t]
\centering
\includegraphics[width=0.9\hsize]{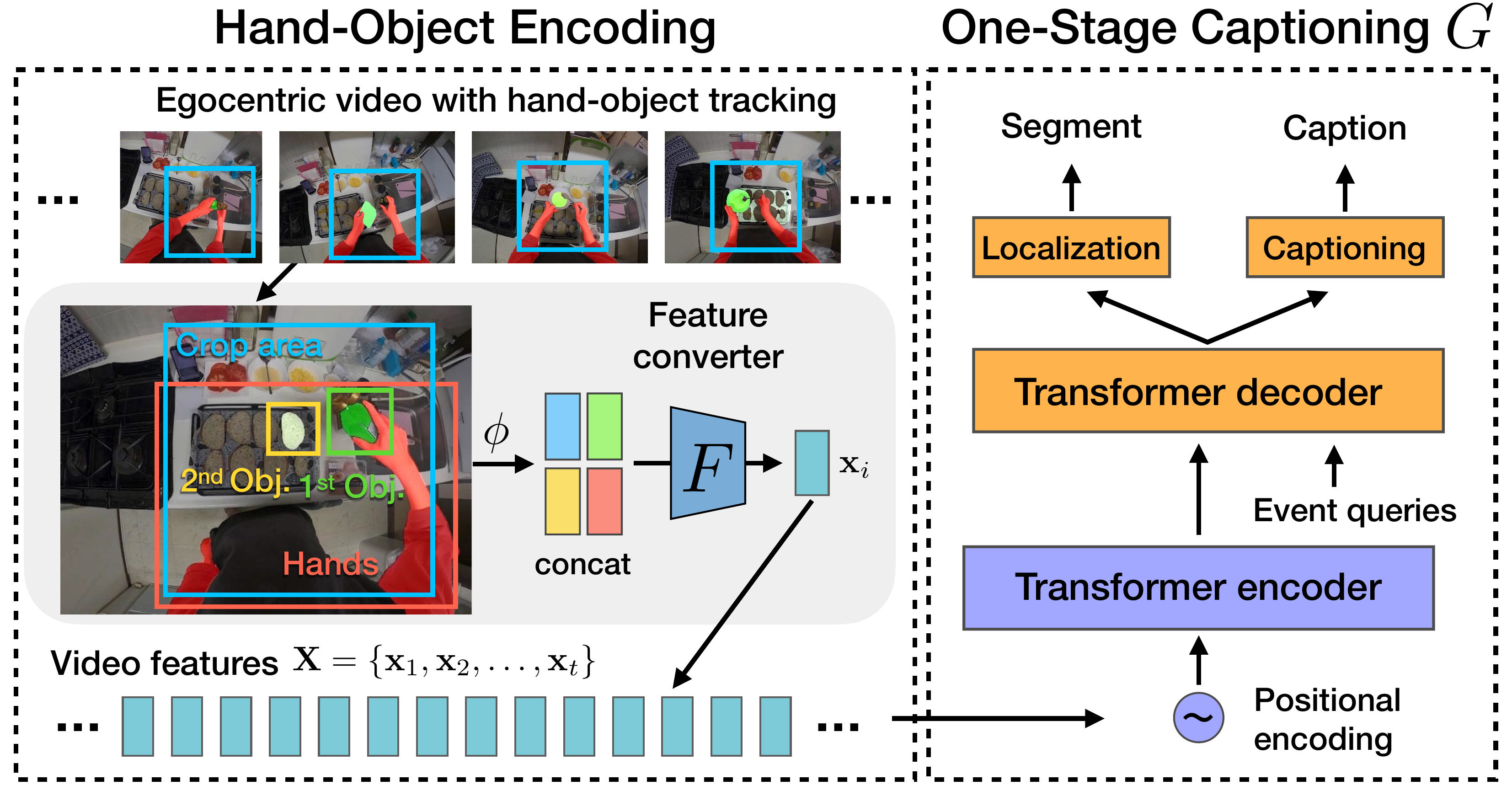}
\caption{
\tbf{Baseline for egocentric dense video captioning.} 
Our baseline consists of (i) hand-object encoding and (ii) one-stage captioning with parallel decoding (PDVC~\cite{wang:iccv21}).
We first preprocess the egocentric videos with hand detection (``crop area'') and hand-object segmentation (``hands'', ``1st obj.'', and ``2nd obj.'').
We extract features for these regions by the fixed encoder $\phi$ and pass their concatenated features to the feature converter $F$.
The generated video features are fed to a transformer-based captioning model with two prediction heads of time segment and caption.
}
\label{fig:baseline}
\end{figure}

\subsection{Hand-object feature generation for model input}\label{sec:hofeat}
We describe feature encoding for egocentric videos.
As discussed in ~\cref{sec:viewproc}, we crop the videos with hand tracking and utilize hand-object masks to localize actions in moving scenes.
Not only reducing the moving effect, these hand and object cues improve recognition of action verbs (\eg, ``pour'') and objects (\eg, ``olive oil'').

Estimates of hand-object interactions have been considered as a means of egocentric video representation, \eg, hand masks~\cite{li:cvpr15}, hand poses~\cite{ohkawa:cvpr23,wen:eccvw24,fan:eccv24}, and 3D hand-object features~\cite{tekin:cvpr19,kwon:iccv21}.
Due to the recent advancement of diverse object segmentation~\cite{kirillov:iccv23}, our work utilizes hand tracking and hand-object segmentation for video representation.

From the affordance analysis~\cite{zhang:eccv22,darkhalil:nips22,yu:wacv23}, hand-object interactions are classified into direct and indirect interactions.
Direct (first-order) interaction refers to the case where the hand is in physical contact with an object. 
Indirect (second-order) interaction describes an object while being manipulated using a tool without direct contact.
In the scene of ``pouring olive oil to bread'' in \cref{fig:baseline}, the hand \textit{directly} grasps an olive oil bottle (\ie, first-order object) and pours it onto bread (\ie, second-order object) where the person \textit{indirectly} interacts with the bread.

\begin{table*}[t]
    \centering    
    \begin{minipage}{0.64\textwidth} 
        \centering
        \caption{
        \tbf{The comparison of datasets for human activity understanding.}
        We show the view type (``ego'' or ``exo'') and whether their views are paired (``P'': paired, ``WP'': weakly paired).
        {We compare the presence of textual annotations and whether they take the form of procedural captions~\cite{zhou:aaai18}.
        }
        The last two columns indicate the domain and the source of the videos.         
        }
        \label{tab:data}
        \resizebox{1.0\linewidth}{!}{
        \begin{tabular}{l|llcccc}
        Dataset                                    & view & \begin{tabular}[c]{@{}c@{}}paired?\end{tabular} & text?                & \begin{tabular}[c]{@{}c@{}}proc.\\ caption?\end{tabular} & domain   & source   \\ \hline
        ActivityNet Captions~\cite{krishna:iccv17} & exo* &                      & {\green{\checkmark}} &                                                          & various  & YouTube  \\        
        YouCook2 (YC2)~\cite{zhou:aaai18}                & exo* &                      & {\green{\checkmark}} & {\green{\checkmark}}                                     & cooking  & YouTube  \\
        EPIC-KITCHENS~\cite{damen:ijcv21}          & ego  &                      & {\green{\checkmark}} &                                                          & cooking  & recorded \\
        MMAC Captions~\cite{nakamura:mm21}         & ego  &                      & {\green{\checkmark}} & {\green{\checkmark}}                          & cooking  & recorded \\
        YouMakeup~\cite{wang:emnlp19}              & exo* &                      & {\green{\checkmark}} & {\green{\checkmark}}                                     & makeup   & YouTube  \\
        COIN~\cite{tang:cvpr19}                    & exo* &                      & {\green{\checkmark}} & {\green{\checkmark}}                                     & various  & YouTube  \\ 
        HowTo100M~\cite{miech:iccv19}              & exo* &                      & {\green{\checkmark}} &                                                          & various  & YouTube  \\
        HIREST~\cite{zala:cvpr23}                  & exo* &                      & {\green{\checkmark}} & {\green{\checkmark}}                                     & various  & YouTube  \\
        Ego4D~\cite{grauman:cvpr22}                & ego  &                      & {\green{\checkmark}} &                                                          & various  & recorded \\ \hdashline
        Charades-Ego~\cite{sigurdsson:cvpr18}      & both & P* &                      &                                                          & various  & recorded \\
        H2O~\cite{kwon:iccv21}                    & both & P &                      &                                                          & various  & recorded \\
        Assembly101~\cite{sener:cvpr22}            & both & P &                      &                                                          & assembly & recorded \\ 
        (3+1)ReC~\cite{sayed:cvprw23}             & both & P &                      &                                                          & cooking  & recorded \\
        Ego-Exo4D~\cite{grauman:cvpr24}           & both & P &                      &                                                          & various & recorded \\ 
        \hdashline
        \tbf{EgoYC2 (Ours)}                       & ego  & WP & {\green{\checkmark}} & {\green{\checkmark}}                                     & cooking  & recorded
        \end{tabular}
        }        
    \end{minipage}
    \hspace{2mm}
    \begin{minipage}{0.3\textwidth} 
        \centering
        \caption{
        \tbf{Statistics of YouCook2 (YC2) and Ego-YouCook2 (EgoYC2).}
        We re-record \dataratio{} of YouCook2 recipes with a head-mounted camera, resulting in \vhour{} hours of \nvideo{} videos.
        }
        \label{tab:egoexo}
        \resizebox{1.0\linewidth}{!}{
        \begin{tabular}{l|cc}
        Datasets               & YC2                & EgoYC2             \\ \hline
        \#video                & 2,000               & 226                \\
        \#user                 & -                   & 44                 \\
        \#recipe\_class        & 89                  & 21                 \\
        total duration         & 176 h              & 43 h               \\
        avg. video dur.        & 5.3 min            & 11.6 min           \\
        avg. segment dur.      & 19 sec             & 103 sec            \\
        avg. step size         & 7.7                & 6.5                \\
        viewpoint              & exo*               & ego                \\
        user consent           & \red{$\times$}     & {\green{\checkmark}}
        \end{tabular}
        }        
    \end{minipage}
\end{table*}

To recognize these objects, we propose a practical refinement scheme of hand-object masks using two segmentation models.
We first adopt EgoHOS~\cite{zhang:eccv22}, a pre-trained segmentation model for hands and interacting objects in egocentric videos.
This model tends to produce inaccurate masks for the objects, as detecting object boundaries is difficult in real-life scenes.
To correct this error, we use a generic segmentation model for any objects (SAM~\cite{kirillov:iccv23}), which can segment objects in any category and generalize to real-world scenes.
Using this capability, we take an intersection between the two models' predictions and use the most overlapped mask from the SAM for the final output.

Given the hand-object masks, we extract features for regions that enclose each entity with the pre-trained encoder $\phi$, namely cropped area (blue), hands (red), first/second-order interacting objects (green/yellow) (see \cref{fig:baseline}).
The feature converter $F$ takes their concatenated features and produces frame-wise features for an entire video.

\subsection{Captioning baseline}\label{sec:videocap}
We employ a one-stage captioning model to construct the task network $G$, which has a unified architecture for the two sub-tasks, predicting time segments (\eg, [033s-125s]) and captions (\eg, ``apply olive oil and cheesy sauce on the bread'').
The primary reason for this choice is that our proposed transfer learning is designed to learn view-invariant features in a middle layer of the encoding pipeline. Thus, using models with separate encoding for the two subtasks would complicate this view-invariant learning.
Specifically, we adopt a strong baseline for dense video captioning with parallel decoding (PDVC)~\cite{wang:iccv21} (see the right of \cref{fig:baseline}).
This model has shown superiority over the two-stage models with the separate encoding scheme~\cite{zhou:cvpr18,wang:tcsvt21}.
The task loss $\mathcal{L}_{\text{task}}$ of PDVC consists of the localization loss of time segments, the classification for event query, the cross-entropy for predicted words, and the cross-entropy for event count.

\section{EgoYC2 Dataset}\label{sec:dataset}
We describe the details of newly collected EgoYC2.
\eccvadd{
To provide a sound evaluation of knowledge transfer between different datasets, EgoYC2 captions follow the caption definition of YouCook2 (YC2)~\cite{zhou:aaai18}.
This ensures that the two datasets are uniform in caption content and granularity, and are evaluated consistently.
Specifically, we directly adopt the \textit{procedural} captions from YC2, describing the sequence of necessary steps to complete complex tasks.
}
We then re-record these cooking videos by instructing participants wearing a head-mounted camera to cook while referring to the YC2's captions (recipes).
We present additional details of the data in the supplement.

\begin{table*}[t]
\begin{minipage}{0.62\textwidth} 
\centering
\caption{
\tbf{
Quantitative results in transfer learning from YouCook2 (YC2) to EgoYC2.
}
We run pre-training (PT) and fine-tuning (FT) with or without the view-invariant (VI) learning.
We also compare various input feature types: raw videos (V), cropped videos (VC), and that with hand-object features (VC+HO).
}
\label{tab:transfer}
\resizebox{1.0\linewidth}{!}{
\small{
\begin{tabular}{ccc|ccc|ccc}
\multicolumn{2}{c}{Method} & \multirow{2}{*}{Input} & \multicolumn{3}{c|}{dvc\_eval}                                                  & \multicolumn{3}{c}{SODA}                                                       \\
\multicolumn{2}{c}{}                      &                        & B4                       & M                        & C                         & M                        & C                        & tIoU       \\ \hline                                   
\hspace{6mm} & Source only  & V & 0.00                     & 0.77                     & 3.6                      & 0.89                     & 1.47                     & 17.9                     \\\hdashline
\parbox[t]{0mm}{\multirow{5}{*}{\rotatebox[origin=c]{90}{Baselines}}} & PT+FT        & V                   & 1.54                     & 7.03                     & 38.1                      & 7.03                     & 25.2                     & 50.5                     \\
&             & VC                  & 1.97                     & 8.20                     & 46.3                      & 8.04                     & 32.3                     & 55.0                     \\
&             & VC+HO                   & 1.68                     & 8.91                     & 52.5                      & 8.91                     & 37.3                     & \underline{59.0}                     \\
& + MMD~\cite{tzeng:arxiv14} & VC+HO & 1.74                     & 8.86                     & 50.9                      & 8.86                     & 37.5                     & 58.8                     \\
& + DANN~\cite{ganin:icml15} & VC+HO & 2.05                     & 9.01                     & 53.1                      & 8.97                     & 39.1                     & 58.6                     \\
\hdashline
\parbox[t]{0mm}{\multirow{3}{*}{\rotatebox[origin=c]{90}{Ours}}} & VI-PT + FT                              & VC+HO & \underline{2.06}                     & \tbf{9.44}                     & \underline{55.2}                      & \underline{9.02}                     & \underline{39.5}                     & 56.0                     \\
& PT + VI-FT                              & VC+HO & 1.77                     & 8.89                     & 53.0                       & 8.91                     & 37.2                     & \tbf{59.1}                     \\
& VI-PT + VI-FT                              & VC+HO & \tbf{2.66}                     & \underline{9.19}                     & \tbf{59.0}                       & \tbf{9.27}                   & \tbf{45.2}                     & 58.1                     \\
\end{tabular}
}
}
\label{tab:transfer}
\end{minipage}
\hspace{2mm}
\begin{minipage}{0.34\textwidth} 
\caption{
\tbf{Analysis of captioning performance with GT proposals.}
We evaluate our comparison models in ~\cref{tab:transfer} given ground-truth (GT) time segments.
We use the VC+HO feature as the input.
``VI'' indicates our proposed method of view-invariant learning introduced in Sec.~\ref{sec:viewinv}. 
}
\label{tab:transfer_gt}
\small{
\resizebox{1.0\linewidth}{!}{
\begin{tabular}{cc|ccc|cc}
\multicolumn{2}{c|}{VI?} & \multicolumn{3}{c|}{dvc\_eval}                                                  & \multicolumn{2}{c}{SODA}                                                       \\
PT                   & FT                   & B4                       & M                        & C                         & M                        & C                                        \\ 
\hline
-                    & -                    & 0.98                     & 10.00                     & 50.6                      & 12.62                     & 40.4                                         \\
\hdashline
{\green{\checkmark}} & -                    & \underline{1.62}                     & \tbf{10.35}                     & \tbf{55.5}                      & \tbf{13.57}                     & \underline{47.9}                                        \\
-                    & {\green{\checkmark}} & 1.07                     & 10.13                     & 51.7                       & 12.70                     & 40.9                                      \\
{\green{\checkmark}} & {\green{\checkmark}} & \tbf{1.68}                     & \underline{10.20}                     & \underline{54.1}                       & \underline{13.36}                   & \tbf{48.7}                                    \\
\end{tabular}
}}
\end{minipage}
\end{table*}

The dataset comparison is shown in ~\cref{tab:data}
\footnote{
``exo*'' view indicates that the video may not be captured from a fixed viewpoint, \eg, YouTube videos contain scene cuts with different views.\\
{\indent ``P*'' in Charades-Ego indicates paired data capturing the same person but not synchronized.}
}.
\eccvadd{ %
Compared to relevant datasets~\cite{sigurdsson:cvpr18,kwon:iccv21,sener:cvpr22,damen:ijcv21,grauman:cvpr22},
our EgoYC2 dataset has unique features in terms of textual annotations and their quality, and pairing to an external dataset.
Instead of using action labels~\cite{sigurdsson:cvpr18,kwon:iccv21,sener:cvpr22}, we newly provide \textit{procedural} captions.
Most ego-exo datasets~\cite{kwon:iccv21,sener:cvpr22,grauman:cvpr24} assume camera synchronization between egocentric and exocentric views, while it is laborious to set up.
In contrast, our benchmark is based on a further relaxed assumption: \textit{weakly paired} setting where different users could perform actions in different recording setups, but the annotated captions follow the same definition between the two datasets.
Furthermore, dense video captioning has not been proposed in popular egocentric video datasets, such as EPIC KITCHENS~\cite{damen:ijcv21} and Ego4D series~\cite{grauman:cvpr22,grauman:cvpr24}, 
and their annotated narrations differ from our \textit{ procedural} captions~\cite{yang:nips23,yang:cvpr23}.
{We compare further details with Ego-Exo4D~\cite{grauman:cvpr24} in the supplement.}
}

\begin{figure}[t]
\centering
\includegraphics[width=0.9\hsize]{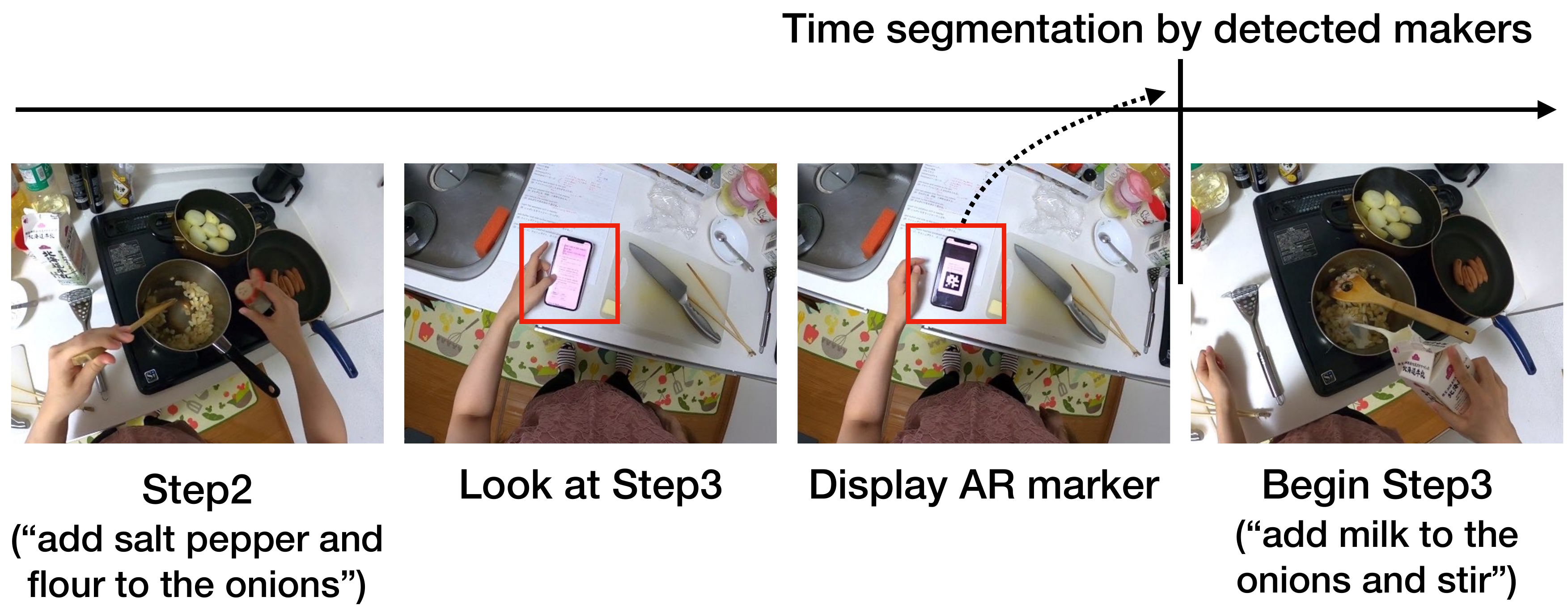}
\caption{
\tbf{Time segmentation by detected AR markers.} In the transition of cooking steps, we ask participants to check the next step on their smartphone or tablet and display an AR marker once they confirm the next step.
Given a recorded video, we postprocess it to detect the marker and segment the video temporally.
}
\label{fig:time}
\end{figure}

Regarding knowledge transfer setups, our benchmark is based on a widely applicable assumption: 
\textit{cross-dataset} transfer, while prior works address the in-dataset transfer, such as Charades-Ego's cross-view training~\cite{sigurdsson:cvpr18} and EPIC-KITCHENS' domain adaptation~\cite{munro:cvpr20}.
This \textit{cross-dataset} assumption does not require strict alignment of the video capture setup between the source and the target data.

\cref{tab:egoexo} shows the statistics compared to YC2.
Our videos reach 11.3~\% of YC2 videos downloaded from YouTube.
Our untrimmed videos have longer video and per-segment duration.
Unlike the MMAC Captions~\cite{nakamura:mm21}, we provide caption annotations following the YC2 to let models address the view gaps without considering caption gaps.
We use the same vocabulary list as YC2 and have close step sizes (7.7 in YC2 \vs 6.5 in EgoYC2 \vs 30.1 in MMAC).
\eccvradd{We obtain user consent for data collection and release.}

\customparagraph{Time segment annotation}
Besides the annotations of step descriptions, we need to annotate the time segment of each step description (\ie, start and end time).
We propose automatic time stamp annotation using AR markers displayed on a virtual screen, as shown in \cref{fig:time}.
We ask participants to use their smartphone or tablet to see a step description in the transition of the steps.
We show an AR marker on the screen once they confirm the next step.
We postprocess a recorded video by detecting the displayed markers, and segment it temporally.
This allows us to annotate the time segments without manually inspecting the entire video.

\section{Experiments}
\subsection{Experimental setup}
\customparagraph{Implementation details}
We employ PyTorch for implementation and run all experiments on a single NVIDIA V100 16GB GPU.
The video features are generated by a pre-trained encoder $\phi$, ResNet152~\cite{he:cvpr16}.
We present different video presentations for egocentric videos: raw video features (V), cropped video features (VC), and the one with hand-object features (VC+HO).
We train models jointly on the source and target data to avoid overfitting the target data.
We set $\lambda_{\text{src}}$ and $\lambda_{\text{adv}}$ as 0.1, and 
{the fixed video length $T$ as 200}.
We denote naive pre-training and fine-tuning as PT+FT and the view-invariant learning as VI-(PT/FT).
Additional details are shown in the supplement.

\customparagraph{Evaluation}
We evaluate the performance in the target domain with five experiments with different random seeds.
We report the average scores of these two metrics: \tbf{dvc\_eval}~\cite{krishna:iccv17} and \tbf{SODA}~\cite{fujita:eccv20}.
The dvc\_eval metric computes the average precision of the matched pairs between the prediction and the ground truth, namely BLEU4 (B4)~\cite{papineni:acl02}, METEOR (M)~\cite{banerjee:aclw05}, and CIDEr (C)~\cite{vedantam:cvpr15}.
The SODA metric considers the storytelling quality for an entire video, \wrt the order of captions and their redundancy.
Following~\cite{nishimura:arxiv22}, we show METEOR (M), CIDEr (C), and temporal Intersection-over-Union (tIoU) as the SODA scores.

\begin{figure*}[t]
\centering
\includegraphics[width=1\hsize]{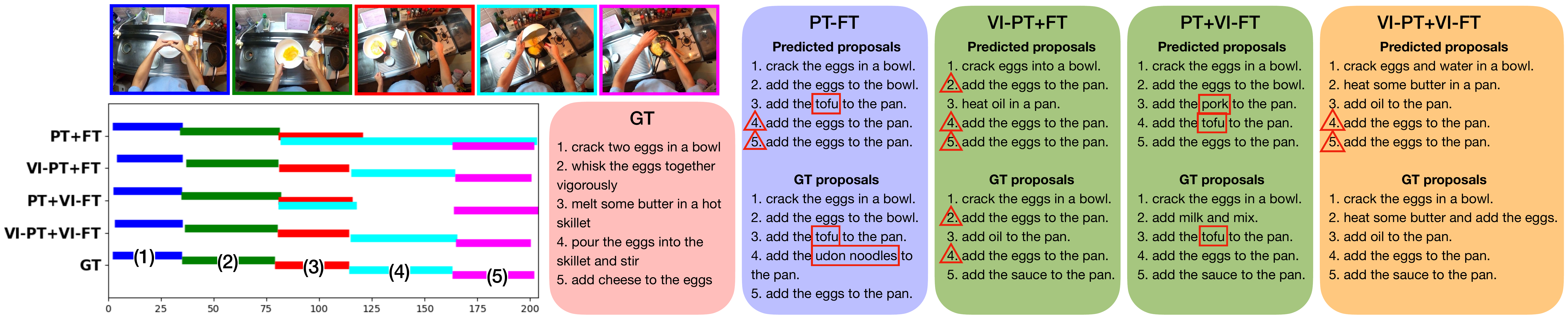}
\caption{
\tbf{Qualitative results} (recipe: scrambled eggs).
We show generated captions given time segment proposals from prediction or ground-truth.
We compare our ablation models: view-invariant (VI) pre-training (PT) and/or view-invariant (VI) fine-tuning (FT).
The marks \red{$\square$} and \red{$\vartriangle$} indicate failure cases for irrelevant ingredients and duplicate captions.
}
\label{fig:caption}
\end{figure*}

\subsection{Results}\label{sec:results}
We report the results of the transfer learning task quantitatively and qualitatively and show the ablation results.
We provide additional results in the supplement, \eg, the choice of hyperparameters, and hand-object mask refinement.

\customparagraph{Results of transfer learning} 
\cref{tab:transfer} shows performance comparisons in the transfer learning setting.
The source only shows the zero-shot generalization ability of a trained model on YC2 to EgoYC2.
The significant view gaps prevent its generalization to egocentric videos.
Our pre-training approach (with view-invariant learning) boosts the performance, suggesting that pre-training on the web video benefits video captioning for egocentric videos.
{
We compare our gradual adaptation with standard domain adaptation methods aligning two-domain features without assuming an intermediate domain, MMD~\cite{tzeng:arxiv14} and DANN~\cite{ganin:icml15}.
Our methods, VI-PT + FT/VI-FT, exhibit better captioning results than those adaptation methods.
This confirms the effectiveness of gradually adapting to the \ego{} view with the guidance of the intermediate domain.
}

\customparagraph{Ablation study of egocentric video representations}
We validate different representations for egocentric videos in \cref{tab:transfer}, namely raw videos (V), cropped videos (VC), and the VC with hand-object features (VC+HO).
We set PT+FT as a base training setting. 
First, a simple cropping technique with hand tracking (VC) is effective, exhibiting a 21.5~\% improvement in the CIDEr score over the V input.
This suggests that tracking the hand region reduces the effect of complex scene changes and makes the task more tractable.
With the hand-object features (VC+HO), we observe the gain of the CIDEr score by 13.4~\% over the VC input and achieve better results. %

\customparagraph{Analysis of captioning capability}
\cref{tab:transfer_gt} shows captioning results given GT time segments without the need for the segment prediction, enabling evaluation of pure captioning ability.
Similarly to \cref{tab:transfer}, models based on view-invariant pre-training (VI-PT + FT and full method) exhibit higher captioning performance.
Since the given time segments will not overlap, the generated captions have fewer duplicate sentences (see \cref{fig:caption}).

\begin{figure}[t]
\centering
\includegraphics[width=0.9\hsize]{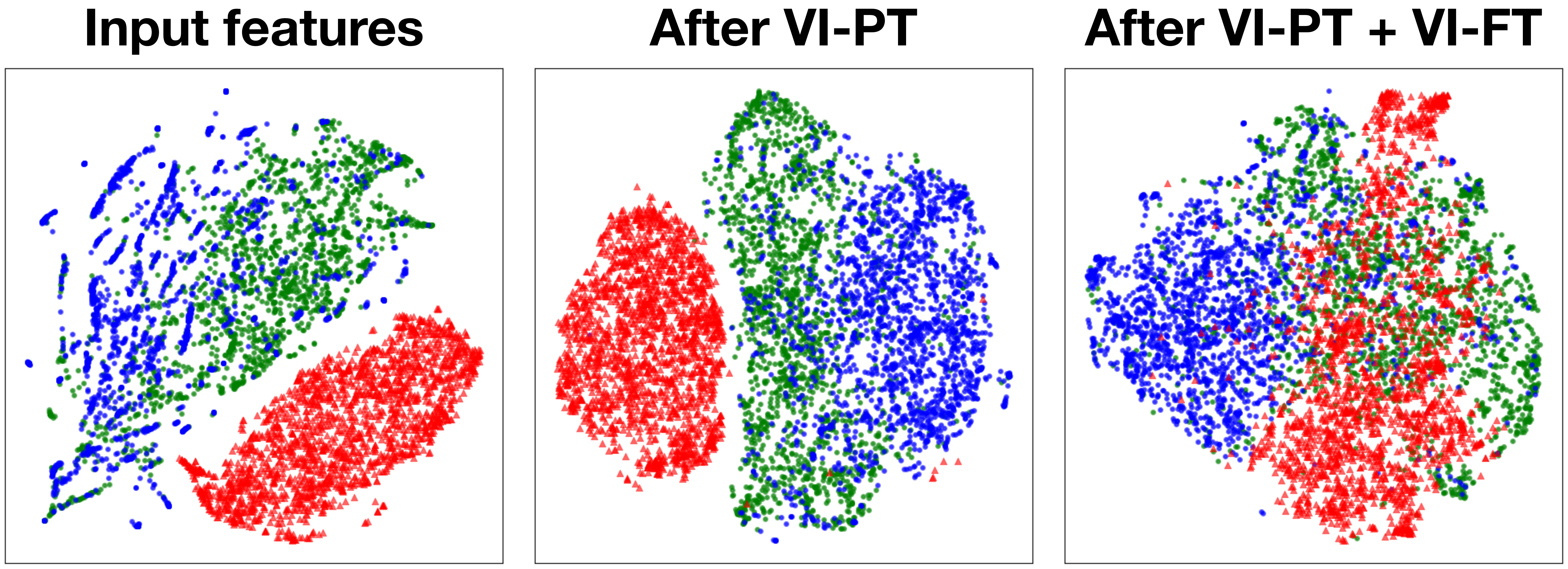}
\captionof{figure}{
\tbf{Visualization of feature distribution}
(\blue{$\CIRCLE$}: \exo{}, {\green{$\CIRCLE$}}: \egolike{}, \red{$\blacktriangle$}: \ego{}).
We visualize the source and target features encoded in each training stage with t-SNE~\cite{maaten:jmlr08}.
Left: initial features generated by the encoder $\phi$, Middle: after view-invariant pre-training (VI-PT) on the source data, Right: after view-invariant fine-tuning (VI-PT + VI-FT) on both datasets.
}
\label{fig:view_dist}
\end{figure}

\customparagraph{Qualitative results}
\cref{fig:caption} shows qualitative results of our ablation models with generated captions and predicted time segments.
For time segmentation, while PT+FT and PT + VI-FT generate overlapped segments and overlook some segments, VI-PT + FT and our full method correct these localization errors.
In generated captions, we observe several failure patterns: appearing unrelated ingredients and duplicate captions.
The captions of PT+FT and PT + VI-FT include the unrelated ingredients (\eg, ``tofu'', ``pork'', and ``udon noodles'').
We also find repeated sentences (triangles) in the models without the view-invariant fine-tuning (\ie, PT+FT and VI-PT + FT).
These observations suggest that the view-invariant pre-training reduces the mixing of unrelated ingredients, and the later view-invariant fine-tuning helps produce fewer repeated sentences.

\customparagraph{Visualization of feature distribution}
To see the transition of the view distribution, we visualize feature distribution for the view classes in \cref{fig:view_dist}.
\eccvadd{%
The visualization is a way to inspect how well the domain gap is mitigated in the feature space~\cite{chen:iccv19,kim:iccv21,munro:cvpr20,ganin:icml15}, which 
indicates that the proximity between the points of different domains represents the model's response to the gap.
}
This confirms that our adaptation method aligns \exo{} and \egolike{} views on the source data first (after VI-PT) and then aligns all three views (after VI-PT + VI-FT).
Interestingly, our method enables aligning visually similar domains with large overlaps between \egolike{} (green) and \ego{} (red) views.

\section{Conclusion}
We present a novel benchmark for cross-view knowledge transfer of dense video captioning from exocentric to egocentric views, together with a new dataset \tbf{\egocap{}}.
We collect \dataratio{} of YouCook2 videos from an egocentric view with aligned captions, enabling transfer learning between both datasets. 
Our proposed view-invariant learning based on adversarial training succeeds in the pre-training and fine-tuning stages while resolving the mixed source views and the moving target view.
We validate our proposed method in the cross-view transfer task with quantitative and qualitative analysis.
This benchmark will promote further studies of transfer learning across the two views and modeling to describe egocentric activities in natural language.

\blfootnote{
\customparagraph{Acknowledgments}
This work was partially supported by JST ACT-X Grant Number JPMJAX2007, JSPS KAKENHI Grant Number JP22KJ0999, JP23H00488, and JP24K02956,  JST AIP Acceleration Research Grant Number JPMJCR20U1, JST ASPIRE Grant Number JPMJAP2303,
and JST Moonshot R\&D Program Grant Number JPMJMS2236, Japan.
This work was also supported in part by a hardware donation from Yu Darvish.
}

\appendix
\section*{Appendix}

\begin{figure}[t]
    \centering
    \includegraphics[width=1\linewidth]{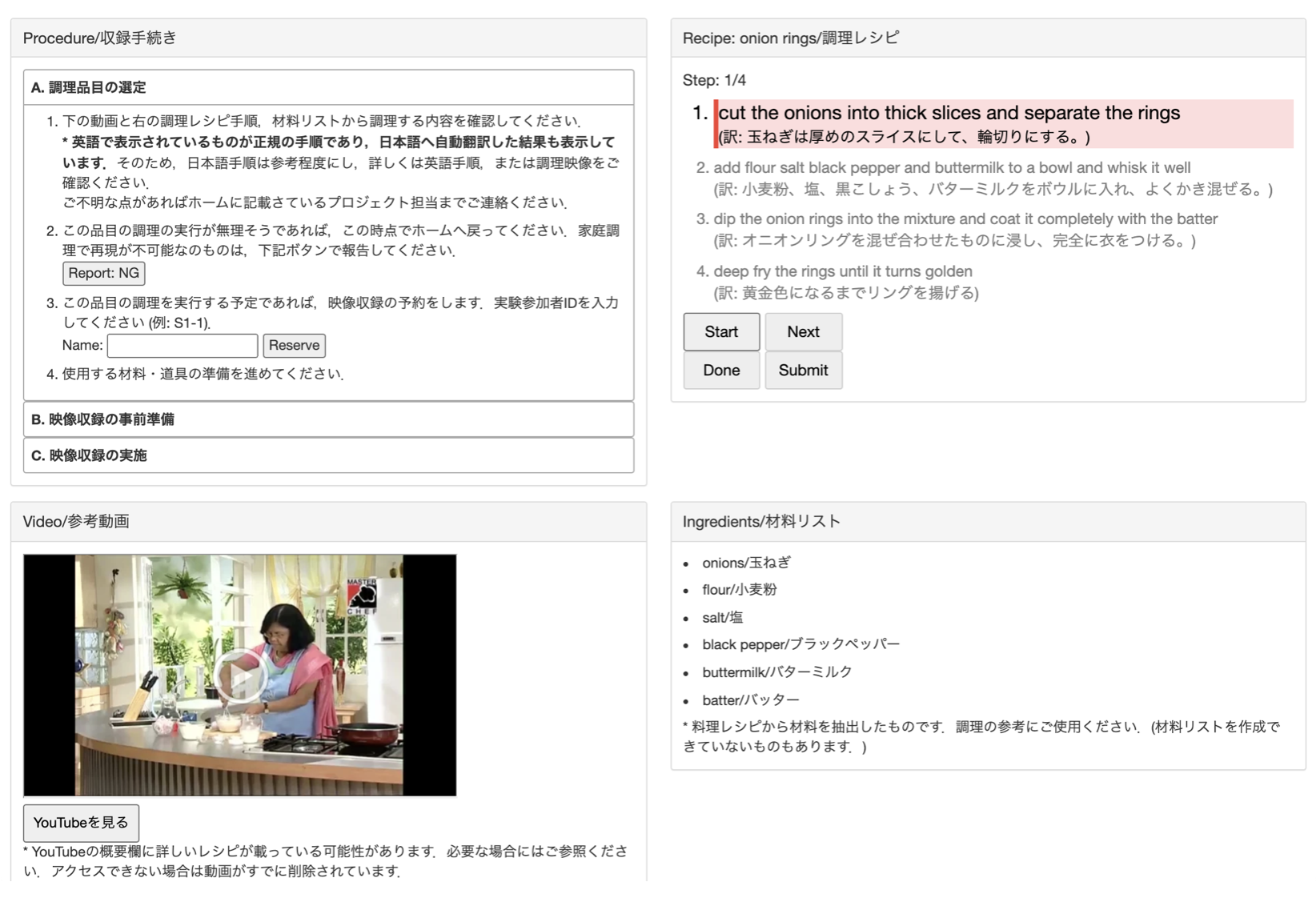}
    \caption{
    \tbf{Web user interface for our recording.}
    Top left: Instruction of recording, Top right: Step description with the focus on the current step, Bottom left: Reference video from YouCook2~\cite{zhou:aaai18}, Bottom right: Necessary ingredients extracted from captions.}
    \label{fig:web_ui}
\end{figure}

\section{Dataset Details}\label{sec:data_details}
\customparagraph{Video recording}
We ask \nuser{} participants to record cooking activities in their own home kitchens using a head-mounted GoPro camera.
The cooking recipes are adopted from YouCook2 (YC2)~\cite{zhou:aaai18} captions with 2,000 recipes consisting of 82 classes of recipes (\eg, ``BLT'' is a class and multiple recipes belong to the class). 
Each participant chooses five recipes at will so that selected classes do not overlap and then prepares the meal by following the step descriptions written in the recipe.
In total, we collect \nvideo{} videos totaling \vhour{} hours.
We also received approval for this activity data collection from an Institutional Review Board and obtained consent from participants who joined this recording.

\cref{fig:web_ui} indicates our web application used for our video recording, displaying the instruction of video collection, step descriptions, reference videos from YouCook2, and necessary ingredients extracted from annotated captions. 
This Web interface helps participants prepare ingredients and check how to cook from the reference videos prior to recording.
During recording, the highlighted step description is shown to indicate their current step and changes by manipulating the button below.
The AR markers are displayed on the screen in the transition of steps, which are used to annotate temporal segments.

\begin{figure}[t]
    \centering
    \includegraphics[width=1\linewidth]{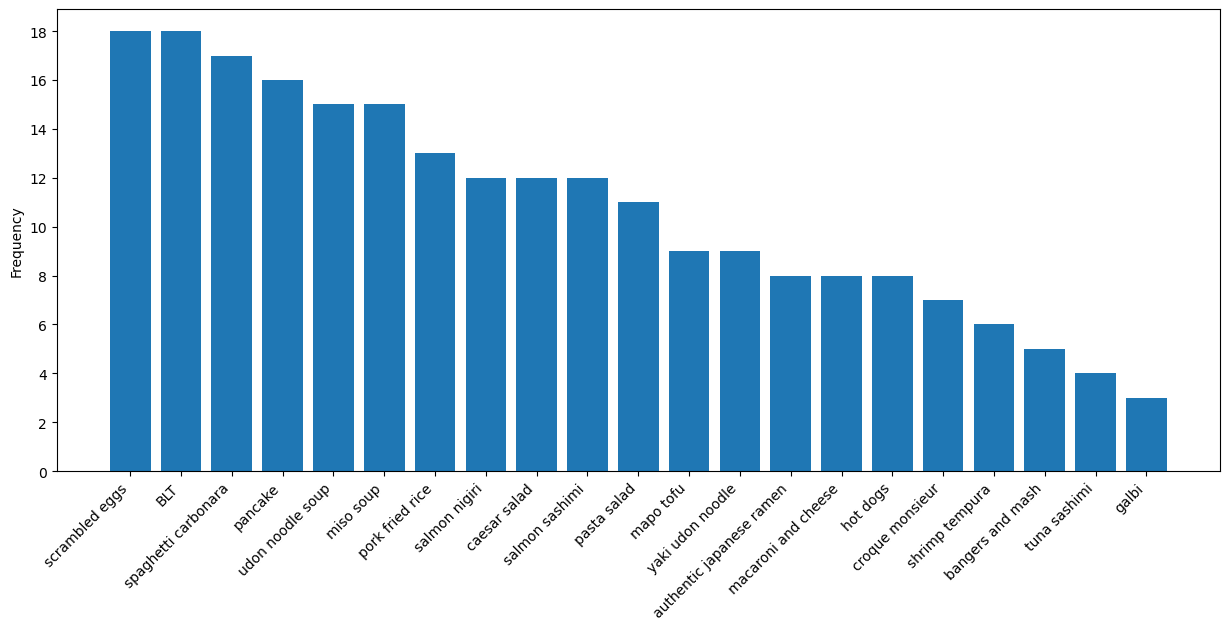} 
    \caption{\tbf{Recipe distribution in EgoYC2}}
    \label{fig:egoyc2_dist}
\end{figure}
\eccvadd{%
To maintain the coherency of captured activities, we instruct the participants to remember the recipes beforehand, which allows them to move to the next step smoothly in the actual recording.
Even though they halted midway through the recording to remember the step procedure, we treat it as acceptable behavior as it is likely to refer to the recipe on their tablets in real-life cooking.
}

\begin{figure*}[h]
    \centering
    \includegraphics[width=1\linewidth]{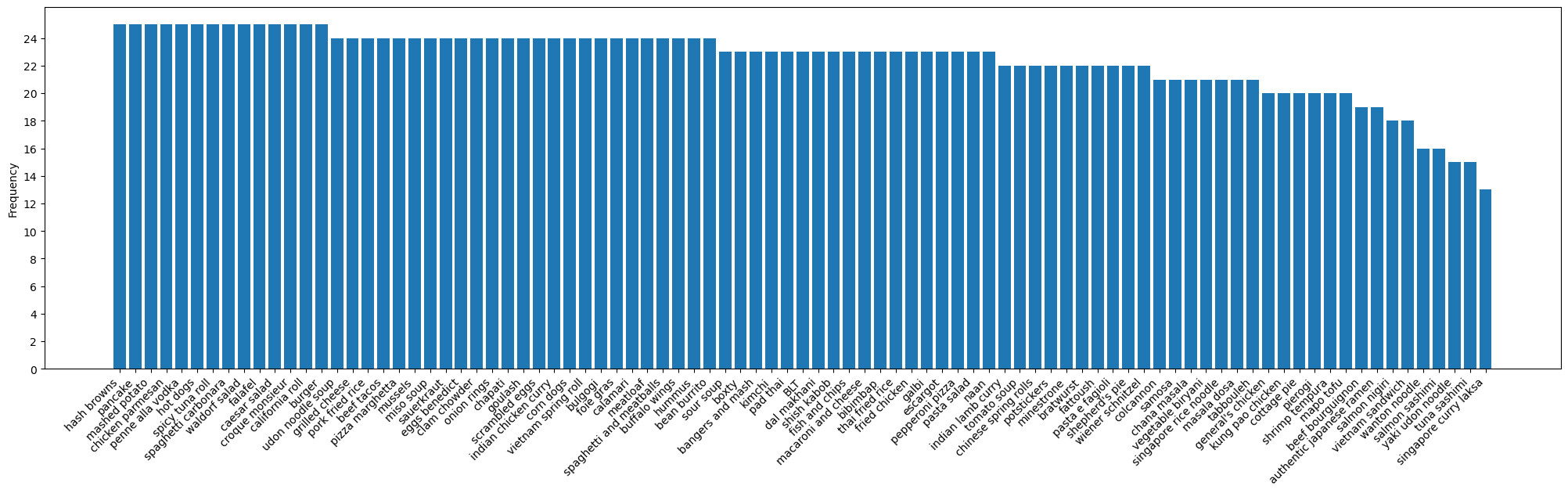} 
    \caption{\tbf{Recipe distribution in YouCook2~\cite{zhou:aaai18}}}
    \label{fig:yc2_dist}
\end{figure*}

\customparagraph{Transfer learning setup}
We use YC2 and EgoYC2 as the source and target data, respectively.
We split the \egocap{} dataset into train and evaluation sets with 151 (964) and 75 (511) videos (step descriptions), respectively. 
To align both datasets, we re-split the YC2 dataset according to the \egocap{}'s split, where train and evaluation sets have 1,716 (13,324) and 75 (511) videos (step descriptions), respectively.
The evaluation sets correspond to each other, and all the YC2 data that are not re-recorded in this work are included in the training set.

\customparagraph{Recipe class distribution}
\cref{fig:egoyc2_dist,fig:yc2_dist} show the distribution of recipe classes for EgoYC2 and YC2.
We collect 21 recipe classes out of 89 classes in YouCook2.
The collected recipe list of EgoYC2 is as follows:  \textit{BLT}, \textit{authentic japanese ramen}, \textit{bangers and mash}, \textit{caesar salad}, \textit{croque monsieur}, \textit{galbi}, \textit{hot dogs}, \textit{macaroni and cheese}, \textit{mapo tofu}, \textit{miso soup}, \textit{pancake}, \textit{pasta salad}, \textit{pork fried rice}, \textit{salmon nigiri}, \textit{salmon sashimi}, \textit{scrambled eggs}, \textit{shrimp tempura}, \textit{spaghetti carbonara}, \textit{tuna sashimi}, \textit{udon noodle soup}, \textit{yaki udon noodle}.

\section{Additional Implementation Details}
The architectures of the feature converter $F$ and the view classifier $C$ follow a two-layer one-dimensional CNN and a three-layer MLP, respectively.
The video features are represented as 2,048-dimensional feature vectors for an input image.
We use PDVC~\cite{wang:cvpr18} as a baseline for dense video captioning.
The PDVC uses a two-layer deformable transformer with a hidden size of 512 in the attention layers and 2,048 in the feed-forward layers. 
The number of event queries is set to 100 and the mini-batch size is set to 1.
We use the Adam~\cite{kingma:iclr14} optimizer with an initial learning rate of 1e-5 for the feature converter and PDVC, and 1e-4 for the view classifier.
While we validate various input types for the target egocentric videos, we use the original video features generated by TSN~\cite{wang:tpami19} on YouCook2.

\section{Additional Results}

\begin{table}
\centering
\caption{
\eccvadd{
\tbf{Quantitative results in scratch training on EgoYC2.}
We train models from scratch in EgoYC2 with various input feature types: raw videos (V), cropped videos (VC), and those with features of an object in hand (VC + HO).
}
}
\label{tab:scratch}
\begin{tabular}{c|ccc|ccc}
\multirow{2}{*}{Input} & \multicolumn{3}{c|}{dvc\_eval}                                                  & \multicolumn{3}{c}{SODA}                                                       \\
                       & B4                       & M                        & C                         & M                        & C                        & tIoU       \\ \hline                                   
V                   & 0.01                     & 3.11                     & 12.3                      & 3.60                     & 5.9                     & 30.7                     \\
VC                  & \textbf{0.10}                     & 5.60                     & 22.2                      & 5.62                     & 12.6                     & 41.5                     \\
VC+HO                   & \textbf{0.10}                     & \textbf{7.34}                     & \textbf{29.6}                      & \textbf{7.04}                     & \textbf{17.9}                     & \textbf{51.4}                    \\
\end{tabular}
\end{table}
\customparagraph{Results with egocentric data only}
\wacvadd{
Tab.~\ref{tab:scratch} shows the results of scratch training on EgoYC2 only.
This demonstrates consistent improvement with hand-object encoding similar to the transfer setup (Tab.~3 in the main paper). %
With paired videos of YC2 (Rows~2-4 in 
Tab.~3 in the main paper),
we observe significant gains over scratch performance, which confirms the effectiveness of transfer learning in limited data regimes for egocentric video captioning.
}

\customparagraph{Analysis of hyperparameter settings}
We set the hyperparameter of view-invariant learning ($\lambda_{\text{adv}}$) by observing the source performance of the view-invariant pre-training (VI-PT).
We use the sum of two METEOR metrics (sum\_METEOR) for the model selection during the pre-training. 
We choose the hyperparameter with the highest sum\_METEOR value and set $\lambda_{\text{adv}}$ as 0.1 consistently for the fine-tuning in the target domain.

We also evaluate performance in the pre-training and fine-tuning stages, according to different hyperparameters in \cref{tab:hp}.
Pre-training with $\lambda_{\text{adv}}=0.01,0.1$ achieves relatively high performance, while fine-tuning with $\lambda_{\text{adv}}=0.01,1$ worsens performance than the PT+FT baseline (top row).
When adding the view-invariant technique to both the pre-training and fine-tuning, we observe an improvement of captioning ability with $\lambda_{\text{adv}}=0.01,0.1$, as they are adapted from the pre-training models where the view-invariant learning performs well.
Based on this study, our setting of $\lambda_{\text{adv}}=0.1$ chosen from the source pre-training performs stably in the target domain with both the pre-training and fine-tuning stages.

\begin{table}[t]
\centering
\caption{
\tbf{Analysis of hyperparameter settings.}
We validate different hyperparameters for the view-invariant learning ($\lambda_{\text{adv}}$) and show the performance on the target dataset.
}
\label{tab:hp}
\small{
\begin{tabular}{llc|ccc|ccc}
\multicolumn{2}{c}{VI?}                         & \multirow{2}{*}{$\lambda_{\text{adv}}$} & \multicolumn{3}{c|}{dvc\_eval}                                                                    & \multicolumn{3}{c}{SODA}                                                                         \\ \cline{4-9} 
\multicolumn{1}{c}{PT} & \multicolumn{1}{c}{FT} &                                           & B4                             & M                              & C                               & M                              & C                              & tIoU                           \\ \hline
                       &                        & 0                                         & 1.68                           & 8.91                           & 52.5                            & 8.91                           & 37.3                           & 59.0                           \\\hline{\green{\checkmark}}   &                        & 0.01                                      & \tbf{2.20}                     & \tbf{9.45}                     & 52.4                            & 8.99                           & \tbf{39.9}                     & 55.0                           \\
{\green{\checkmark}}   &                        & 0.1                                       & 2.06                           & 9.44                           & \tbf{55.2}                      & \tbf{9.02}                     & 39.5                           & \tbf{56.0}                     \\
{\green{\checkmark}}   &                        & 1                                         & 1.70                           & 9.29                           & 50.5                            & 8.75                           & 36.4                           & 55.1                           \\\hdashline
                       & {\green{\checkmark}}   & 0.01                                      & 1.47                           & 8.75                           & 49.8                            & 8.72                           & 35.8                           & 58.8                           \\
                       & {\green{\checkmark}}   & 0.1                                       & \tbf{1.77}                     & \tbf{8.89}                     & \tbf{53.0}                      & \tbf{8.91}                     & \tbf{37.2}                     & 59.1                           \\
                       & {\green{\checkmark}}   & 1                                         & 1.50                           & 8.67                           & 49.4                            & 8.67                           & 35.6                           & \tbf{59.5}                     \\\hdashline{\green{\checkmark}}   & {\green{\checkmark}}   & 0.01                                      & \multicolumn{1}{l}{2.46}       & \multicolumn{1}{l}{\tbf{9.60}} & \multicolumn{1}{l|}{53.1}       & \multicolumn{1}{l}{8.99}       & \multicolumn{1}{l}{39.3}       & \multicolumn{1}{l}{55.4}       \\
{\green{\checkmark}}   & {\green{\checkmark}}   & 0.1                                       & \multicolumn{1}{l}{\tbf{2.66}} & \multicolumn{1}{l}{9.19}       & \multicolumn{1}{l|}{\tbf{59.0}} & \multicolumn{1}{l}{\tbf{9.27}} & \multicolumn{1}{l}{\tbf{45.2}} & \multicolumn{1}{l}{\tbf{58.1}} \\
{\green{\checkmark}}   & {\green{\checkmark}}   & 1                                         & \multicolumn{1}{l}{1.58}       & \multicolumn{1}{l}{9.30}       & \multicolumn{1}{l|}{49.7}       & \multicolumn{1}{l}{8.67}       & \multicolumn{1}{l}{35.3}       & \multicolumn{1}{l}{54.9}      
\end{tabular}
}
\end{table}

\begin{figure*}[t]
    \centering
    \includegraphics[width=1\linewidth]{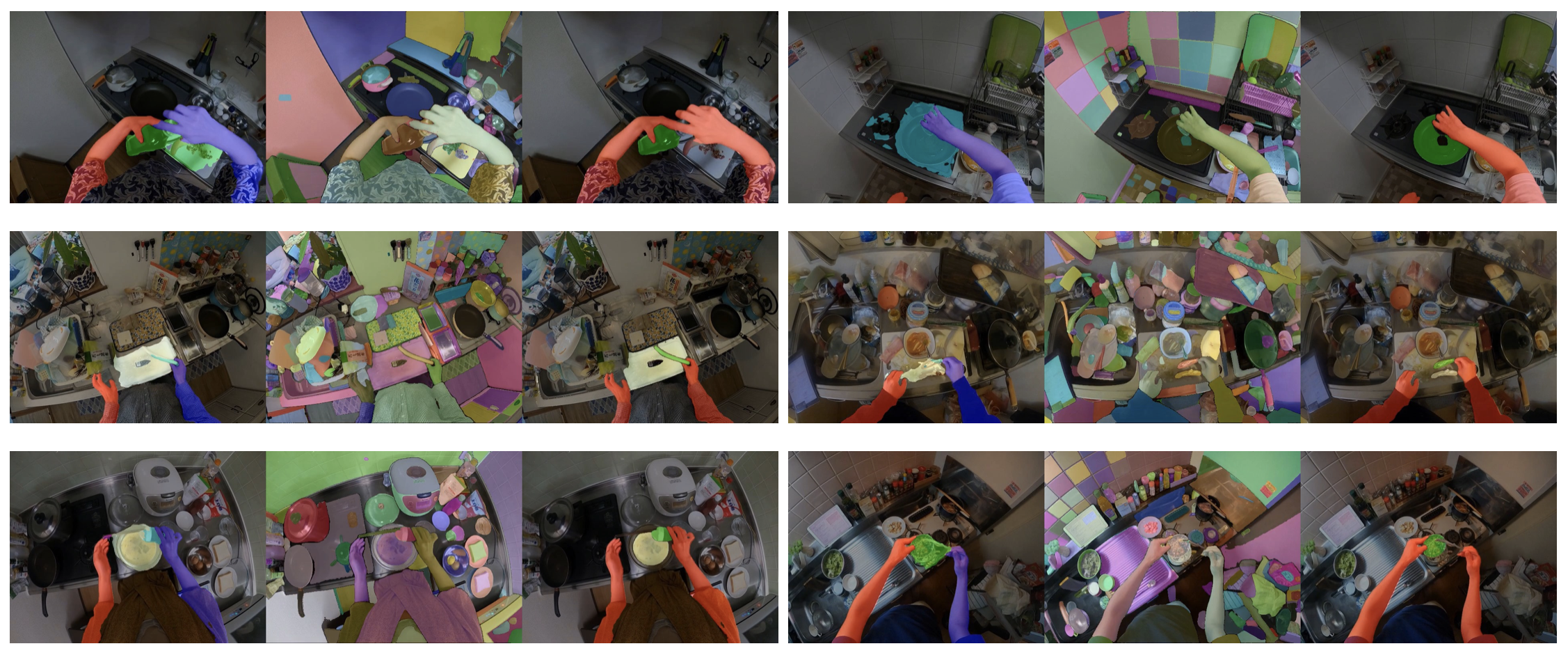}
    \caption{\tbf{Our hand-object segmentation refinement.}
    Each panel shows segmentation results of EgoHOS~\cite{zhang:eccv22} (left), SAM~\cite{kirillov:iccv23} (middle), and our refined scheme (right), respectively. Since we don't use hand identity information (right/left), we show merged hand masks compared to the results of EgoHOS.
    }
    \label{fig:ho_segm}
\end{figure*}

\customparagraph{Hand-object segmentation results}
We propose a segmentation refinement scheme based on two segmentation models: EgoHOS~\cite{zhang:eccv22} and SAM~\cite{kirillov:iccv23}.
We show the segmentation results for each method in \cref{fig:ho_segm}.
The EgoHOS inference (left) often has noisy results (\eg, undersegmetation on the top row and incorrect localization of long and narrow objects on the middle row).
EgoHOS suffers from generalizing to novel real-life environments where diverse object types and shapes could be present.
The SAM inference (middle) can segment any kind of object with higher generalization.
Our refinement (right) computes the overlap between the two results and outputs the most overlapped segments from the SAM predictions.
This enables us to obtain further refined results even in crowded cooking environments (\eg, middle row).

\section{Discussions}
\eccvradd{
\customparagraph{Scripted \vs unscripted}
Scripted and unscripted captures each have pros and cons concerning data realism and annotation quality.
While unscripted videos, such as Ego4D~\cite{grauman:cvpr22} and EPIC-KITCHENS~\cite{damen:ijcv21}, reflect actual activities, these videos could include ambiguity in captions from human annotators, affecting the consistency of caption content and granularity.
Such inconsistency complicates cross-domain evaluation. 
Our scripted approach not only aligned the content and granularity between datasets, but also instructed participants to maintain action coherency
in Sec.~\ref{sec:data_details},
enabling natural step transitions in captured videos.
}

\eccvradd{
\customparagraph{Unsupervised methods}
Zero-shot generalization and unsupervised adaptation remain challenging in video captioning, as evidenced by the source-only results shown in Tab.~3 of the main paper.
Our benchmark provides supervised baselines and evaluations on egocentric videos, setting the stage for future studies to develop unsupervised methods. 
}

\customparagraph{Overcoming recipe class gap}
As shown in Sec.~\ref{sec:data_details}, the recipe class distribution is not perfectly aligned between YC2 and EgoYC2. 
In addition to focusing on the view gap addressed in the main paper, resolving category shift~\cite{cao:eccv18,luo:cvpr19,xu2:iccv21}, the gap in the output (label) space, will be an important future challenge.

\customparagraph{
{Comparison with Ego-Exo4D}}
{
We provide the comparison with a recently released Ego-Exo4D dataset~\cite{grauman:cvpr24}, featuring synchronized egocentric and exocentric videos with textual annotations. 
In capture setups, the work follows the strong assumption of time-synchronized and calibrated scenarios, while our captures between YC2 and EgoYC2 are based on a relaxed assumption; they are not synchronized and not captured in the same environment.
Regarding its text annotations~\footnote[1]{\href{https://docs.ego-exo4d-data.org/annotations/atomic_descriptions/}{https://docs.ego-exo4d-data.org/annotations/atomic\_descriptions/}}, the knowledge of the coherency between steps is not explicitly modeled, as each description is instructed to be annotated independently.
In contrast, our procedural captions are intended to model the necessary steps to accomplish a target task, which inherently includes inter-step relationships in the captions.
}

{\small
\bibliographystyle{cfgs/ieee_fullname}
\bibliography{241128_main_purified.bbl}
}

\end{document}